%% file: main.tex
\documentclass[10pt,twocolumn,letterpaper]{article}

\usepackage{iccv}
\usepackage{times}
\usepackage{epsfig}
\usepackage{graphicx}
\usepackage{amsmath}
\usepackage{amssymb}

\usepackage{xspace}
\usepackage{color}
\usepackage{booktabs}
\usepackage{subcaption}
\usepackage{bbm}
\usepackage{microtype}

\usepackage{algorithm}
\usepackage{algorithmic}

\usepackage[pagebackref=true,breaklinks=true,letterpaper=true,colorlinks,bookmarks=false]{hyperref}

\iccvfinalcopy % *** Uncomment this line for the final submission

\ificcvfinal\fi

\begin{document}

%%%%%%%%% TITLE
\title{The Story in Your Eyes:\\ An Individual-difference-aware Model for Cross-person Gaze Estimation}

\author{Jun Bao\\
Hangzhou Dianzi University\\
Hangzhou, China\\
{\tt\small baoj@hdu.edu.cn}
\and
Buyu Liu\\
NEC Laboratories America\\
San Jose, USA\\
{\tt\small buyu@nec-labs.com}
\and
Jun Yu\\
Hangzhou Dianzi University\\
Hangzhou, China\\
{\tt\small yujun@hdu.edu.cn}
}

\maketitle

\ificcvfinal\thispagestyle{empty}\fi

%%%%%%%%% BODY TEXT
\input{sections/abstract.tex}
\input{sections/intro.tex}
\input{sections/related_work.tex}
\input{sections/method.tex}
\input{sections/experiments.tex}
\input{sections/conclusion.tex}

{\small
\bibliographystyle{ieee_fullname}
\bibliography{main}
}

%-------------------------------------------------------------------------
%\newpage
%\input{supp_material.tex}
%{\small
%\bibliographystyle{ieee_fullname}
%\bibliography{egbib_supp}
%}

\end{document}

%% file: sections/abstract.tex
%%%%%%%%% ABSTRACT
\begin{abstract}
We propose a novel method on refining cross-person gaze prediction task with eye/face images only by explicitly modelling the person-specific differences. Specifically, we first assume that we can obtain some initial gaze prediction results with existing method, which we refer to as InitNet, and then introduce three modules, the Validity Module (VM), Self-Calibration (SC) and Person-specific Transform (PT)) Module. By predicting the reliability of current eye/face images, our VM is able to identify invalid samples, e.g. eye blinking images, and reduce their effects in our modelling process. Our SC and PT module then learn to compensate for the differences on valid samples only. The former models the translation offsets by bridging the gap between initial predictions and dataset-wise distribution. And the later learns more general person-specific transformation by incorporating the information from existing initial predictions of the same person. We validate our ideas on three publicly available datasets, EVE, XGaze and MPIIGaze and demonstrate that our proposed method outperforms the SOTA methods significantly on all of them, e.g. respectively $21.7\%$, $36.0\%$ and $32.9\%$ relative performance improvements. We won the GAZE 2021 Competetion on the EVE dataset. Our code can be found here {\small{\url{https://github.com/bjj9/EVE_SCPT}}}.
\end{abstract}

%% file: sections/intro.tex
\section{Introduction}
\label{sec:intro}
Gaze estimation from a single low-cost RGB sensor is an important topic in computer vision~\cite{Park2020ECCV,park2019few}. Inputting face or eye images~\footnote{Our method can handle both eye and face images and we refer to our input later in this paper as eye images for simplicity.}, gaze prediction aims to estimate the gaze direction or Point of Gaze (PoG) on screen. The task of cross-person gaze estimation is defined as one where a model is evaluated on a previously unseen set of participants, which is undoubtedly more challenging. Results from gaze prediction usually serves as input to tasks, e.g. human attention estimation~\cite{Chong2018ConnectingGS}. Therefore, having an accurate gaze estimation model can be essential for downstream tasks, e.g. intelligent user interfaces~\cite{Feit_17,Huang_16}.% and user state awareness.

Though there are many existing work aim to predict gaze location/direction w.r.t information from single images~\cite{park2019few}, video sequences and contents~\cite{Park2020ECCV}, the problem that there exist un-observable person-specific differences remains. To this end, many person-specific adaptation techniques are proposed, including exploiting networks to predict person-wise differences~\cite{liu2019differential}, directly estimating 6-degree calibration parameters of human eyes~\cite{linden2019learning} or fine-tuning models with few labelled test samples~\cite{yu2019improving}.

\begin{figure}[t]\centering
  \includegraphics[width=1.0\columnwidth]{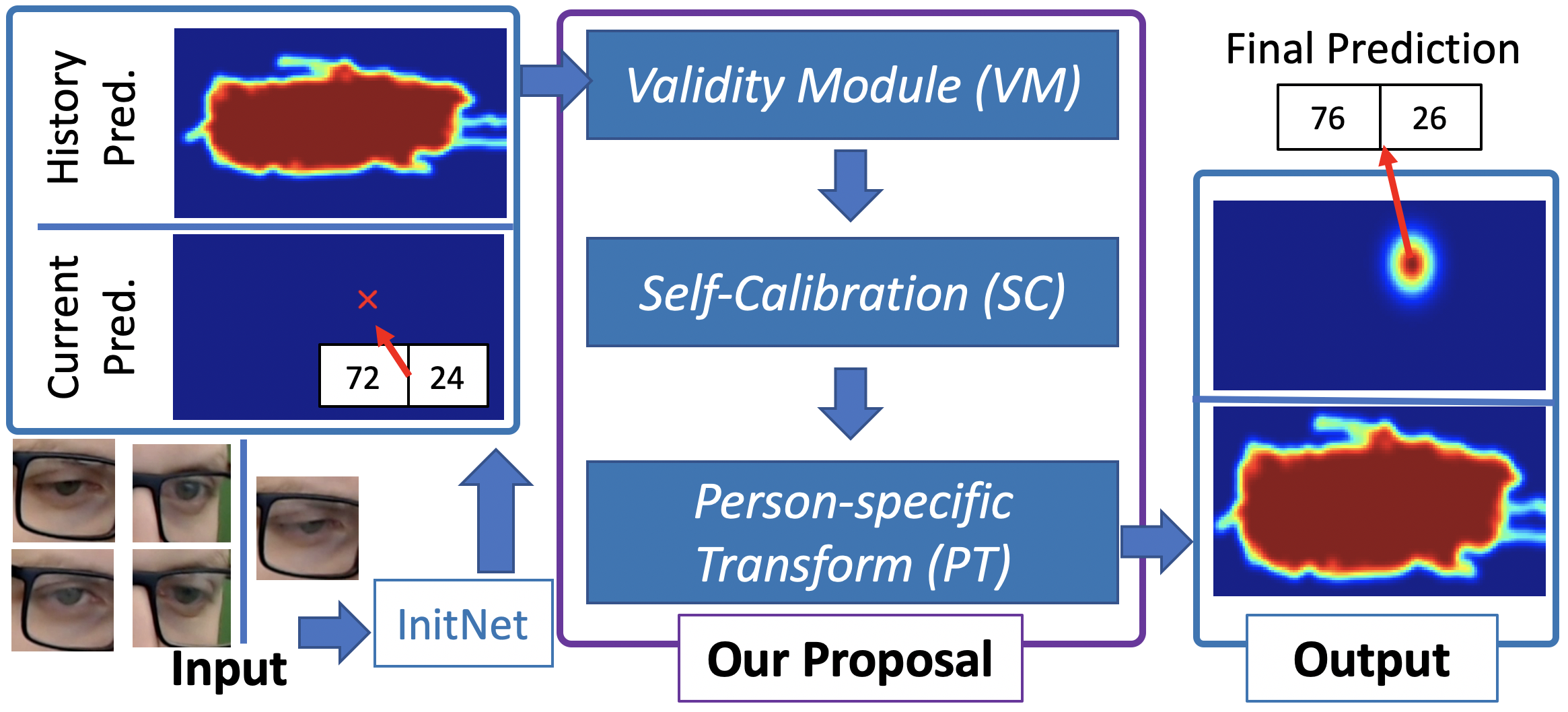}
  %\fbox{\rule{0pt}{2in} \rule{.9\linewidth}{0pt}}
  % \vspace{-0.65cm}
  \caption{
  We propose to model the person-specific differences in cross-person gaze estimation problem. Specifically, our model takes the initial gaze predictions of current eye image and previous initial predictions of the same participant as input. After removing invalid samples via VM and explicitly modelling the person-specific offsets, it is able to improve the gaze prediction accuracy by a large margin.
  }
  \label{fig:teaser}
\end{figure}

We follow this adaptation line of work and propose a novel method to model person-specific differences. Unlike existing methods that require additional information, such as labelled samples from test participants to perform person-specific calibration or additional content information on screen, ours learns to compensate person-specific differences with eye images \textbf{only}. To achieve that, we propose three modules, one for modelling the reliability of samples and two others for person-specific differences modelling. Specifically, we introduce a non-parametric module, or Validity Module (VM), that estimates the reliability of samples so that the invalid/unreliable samples would not affect our difference modelling procedure. We further propose two modules, Self-Calibration (SC) and Person-specific Transform (PT), to bridge the person-specific differences between human optical and visual axes. Our SC Module takes initial prediction from existing methods as input, models the gap between initial prediction w.r.t. dataset-wise gaze distribution and outputs refined prediction. And the PT Module explicitly learns a more general transformation on refined predictions %and ground-truth prediction while 
by incorporating information from history refined predictions of the same participant. Please note that history predictions of the same participant can be interpreted in two ways. In the online setting where we are given consecutive frames as input, %and are requested to make predictions on the fly, 
history predictions of the same participant mean all previous predictions from this particular participant. Given an offline scenario where all eye images, either sampled sparsely in video sequences or just some random images, of one participant are available, we then refer to all other images except the current one as its history. In general, our method is applicable to both video and image based dataset as long as %more than two samples 
multiple samples are available for one participant during test time. %And our method is generic for both image or video based data.

% motivation
It is well-acknowledged in vision literature that the angle kappa, the deviation between optical axis and visual axis, cannot be reflected by images taken with conventional cameras~\cite{basmak2007measurement}~\footnote{It can be captured with dedicated equipment, e.g. Synoptophore.}. The standard deviation of angle kappa in normal population is around 1.8$^{\circ}$ in both horizontal and vertical axis~\cite{berrio2010optical, atchison2000optics}, leading to a random error of 2.0$^{\circ}$-2.3$^{\circ}$ when estimating gaze directions directly from images. 
%Based on the findings above, 
Therefore, existing single-image-based methods have no chance to recover such individual differences and, in theory, their accuracy would ceil at 2.0$^{\circ}$. Our goal is to model such differences thus effectively close the gap.

We validate our ideas on three publicly available datasets, EVE~\cite{Park2020ECCV}, MPIIGaze~\cite{zhang2017mpiigaze} and XGaze~\cite{zhang2020eth}, and obtain the state-of-the-art (SOTA) performance on all three datasets,  $27.6\%$, $36.0\%$ and $32.9\%$ relative performance improvement over existing SOTA methods. We further demonstrate the effectiveness of each module by performing ablation studies.% on EVE. 

To summarize, our key contributions are:
\begin{itemize}
\item A \textbf{novel} gaze estimation method handles data with invalid samples and models person-specific differences under cross-person setting with only eye images. %without acquiring any labelled ground-truth from test participants or additional information other than eye images.
\item Our method includes (i) a Validity Module that estimates the reliability of samples, (ii) a Self-Calibration Module that models prediction offset w.r.t. dataset-wise distribution and (iii) a Person-specific Transform Module that explicitly compensates individual differences.
\item Our model outperforms the SOTA by a large margin on three publicly available datasets. We won the GAZE 2021 Competetion on the EVE dataset.
\end{itemize}

%% file: sections/related_work.tex
\section{Related Work}
\label{sec:related}
\paragraph{Model-based Gaze Estimation} Gaze estimation methods can generally be categorized into model-based or appearance-based. Model-based methods generally exploit geometric eye models and can be further distinguished into shape-based and corneal-reflection methods. Shape-based methods~\cite{ishikawa2004passive,chen20083d} first detect eye shape and then infer gaze directions from detection results, e.g. the pupil centers. In contrast, corneal-reflection models rely on eye features by using reflections of an external infrared light source on the outermost layer of the eye, the cornea. Starting from limited stationary settings~\cite{morimoto2002detecting}, corneal-reflection are capable of handling arbitrary head poses using multiple light sources or cameras~\cite{zhu2005eye} nowadays. Though more practical application scenarios~\cite{cristina2016model,wood2014eyetab} have been observed with model-based methods, one main drawback of this line of work remains. That is, their gaze estimation accuracy is not satisfactory for real-world settings as they rely heavily on accurate eye feature detection results, which further requires high-resolution images and homogeneous illumination. Due to such requirements, these methods are not as desired as appearance-based methods in real-world settings or on commodity devices.

\paragraph{Appearance-based Gaze Estimation} Appearance-based gaze estimation methods aim to map images directly to gaze. Compared to model-based approaches, appearance-based method achieves much better results for in-the-wild settings. Intuitively, these methods do not rely on any outputs from explicit shape extraction step thus are less constrained to image resolution or distances. Under restricted conditions, e.g. images are taken under limited and constrained laboratory conditions, regression techniques~\cite{lu2011inferring} or random forests ones~\cite{sugano2014learning} have been explored. %under appearance-based methods. 
With the benefits of large scale datasets, e.g. MPIIGaze~\cite{zhang2017mpiigaze}, CNN-based methods~\cite{zhang2015appearance,zhang2017mpiigaze} further push this field fast forward. MPIIGaze therefore becomes a benchmark dataset for in-the-wild gaze estimation. Recently, larger datasets, such as EVE~\cite{Park2020ECCV} and XGaze~\cite{zhang2020eth}, provide more diverse data to evaluate gaze prediction performances with various experimental settings. To improve the prediction performance, researchers have explored model structures, e.g. introduce more complex or ensembles of CNNs~\cite{zhang2017mpiigaze,fischer2018rt}, model input, such as extend to multi-modal input~\cite{krafka2016eye,yu2018deep} or improve data normalization~\cite{zhang2018revisiting}, and model representations, for instance, learning more informed intermediate representations~\cite{park2018deep}. However, in order to cover the significant variability in eye appearance caused by free head motion, these methods require more person-specific training data compared to model-based approaches, %. One of the limitations of traditional appearance-based methods is that 
e.g. they generally require some specific domains or persons. Compared to existing methods, ours is more generic. %in two folds. 
Firstly, all the above mentioned methods, either appearance-based or model-based, are compatible with our model as long as they provide initial gaze estimations and our method can be applied to improve their performance. Secondly, our method %is able to 
explicitly models person-specific differences with eye images only.%.with neither seeing any labelled samples of test participants nor requiring additional information other than eye images.

\paragraph{Cross-person Gaze Estimation} Modelling person-specific differences seems to be a natural way to perform cross-person estimation task. However, given restricted setting that the test participants are unseen during training time, incorporating personal modelling/calibration can be a challenging problem itself. Therefore, some existing methods try to explore additional data to tackle the cross-person gaze direction (and subsequent Point-of-Gaze) prediction problem. %To this end, various methods have been proposed. 
One line of work proposed to relax the restriction a bit by assuming that very few samples of a target test person’s data are available. Then they fine-tuned or adapted pre-trained model on this data and demonstrate advanced performance on the final test data from the same person~\cite{krafka2016eye,park2018learning}. Building on top of this work, less samples, e.g. as few as 9 calibration samples for each test person, are required to achieve performance improvements~\cite{liu2019differential}. Although the improvements seem to be promising, these methods all require labeled samples of test participants thus is not practical/generic. Another type of additional information comes from screen contents where the predicted visual saliency of the screen content is assumed to be available. With this assumption, researchers propose to align estimated PoG with an estimated visual saliency~\cite{sugano2010calibration,chen2011probabilistic}. To avoid the over-fitting problem of single saliency models on training data, multiple saliency models~\cite{sugano2015self} are explored. More recent work~\cite{zhang2020eth} exploit both screen content and temporal cues. Unlike saliency-based method, screen content is incorporated in the form of region of interests on screen~\cite{zhang2020eth}. In contrast to existing methods that require additional information for person-specific modelling, such as test samples to be labelled, screen content or even consecutive frames from video sequences, our proposed method is more generic since we requires eye images~\textbf{only}. 

%% file: sections/method.tex
\section{Our Framework}
\label{sec:method}
Our model includes four modules. The first module InitNet inputs eye images and outputs the initial Point of Gaze (PoG) on screen. The second Validity Module (VM) inputs the initial PoG predictions and outputs the reliability of each sample. The third module, or Self-Calibration (SC) Module, takes the validity predictions as well as the initial predictions as input, and learns to compensate the current prediction w.r.t. its history predictions and dataset-wise distribution. And we refer to the output from SC as refined PoG. Our last Person-specific Transform (PS) Module parses the refined PoG and its refined history information % in the form of heatmaps and 
outputs our final PoG by explicitly modelling person-specific differences. We provide more details for each module in Sec.~\ref{sec:model_model} and then introduce learning process of each module in Sec.~\ref{sec:model_train}. Please note that our main contributions lie in the last three modules and we introduce the first module here for %paper 
completeness. 

\begin{figure*}\centering
  \includegraphics[width=1.0\linewidth]{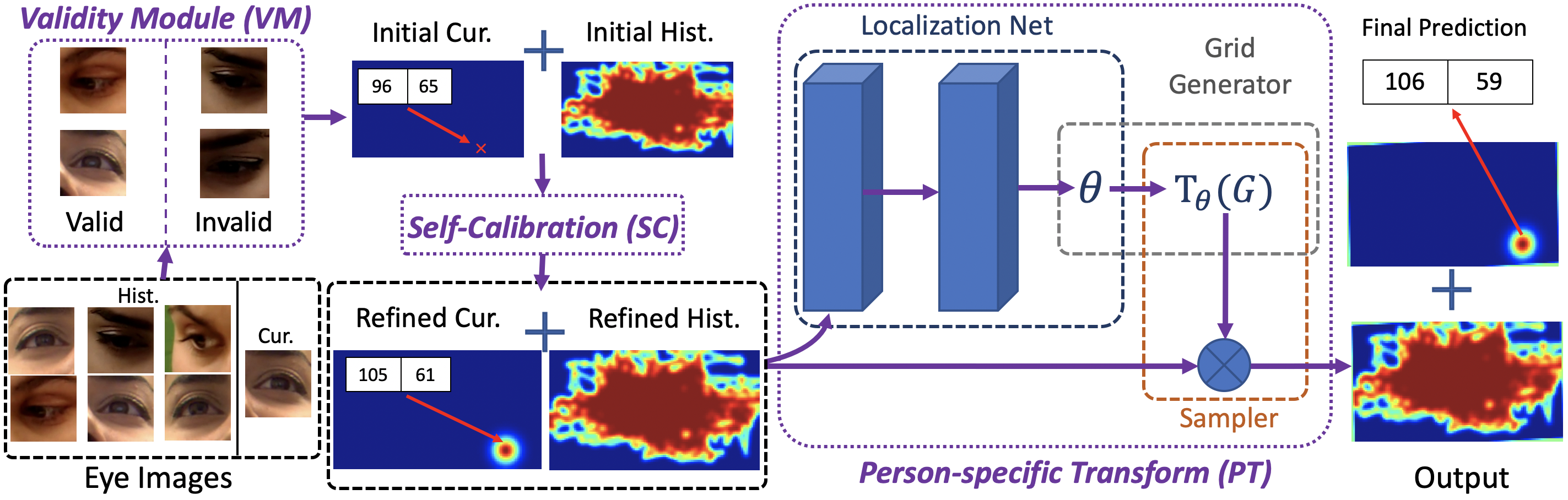}
  %\fbox{\rule{0pt}{2.7in} \rule{.9\linewidth}{0pt}}
  \caption{
  Assuming that we have current eye image and history eye images of the same person available, our method first models the validity of all images with the help of VM and then explicitly compensates the person-specific differences with SC and PT modules. We observe significant performance improvement over cross-person gaze estimation task.
  }
  \label{fig:main}
\end{figure*}

\subsection{Our Model}~\label{sec:model_model}
In this section, we focus on model structure and assume that supervisions are available for each module. Let's first assume that we have an eye-tracking data set$~\mathcal{D}=\{I_{t,j}, g_{t,j}, v_{t,j}\}_{t=1,j=1}^{N,M}$ of $N$ samples for $M$ participants. Specifically, $I_{t,j} =\{I_{t,j}^*\}_{*={l,r}}$ representing the paired images for the $t$-th sample of $j$-th participants, where $*=l$ represents the left eye and $*=r$ is for the right. We define $g_{t,j}$ as a 2-dimensional vector denoting the ground truth Point of Gaze (PoG) location of the $t$-th sample of the $j$-th participant on screen. Finally, $v_{t,j}$ is binary indicating whether the current sample is valid or not. Fig.~\ref{fig:main} gives an overview of the proposed method.

~\subsubsection{InitNet Module}~\label{sec:eyenet}
The goal of InitNet module is to provide initial PoG estimation on the given images. Please note that the ultimate goal of this paper is to improve the gaze predictions from InitNet %thus InitNet is not our contribution 
and it can be any existing gaze estimation methods. %of any form as long as it provides initial PoG predictions. 
We denote the initial PoG prediction of the $t$-th paired images from participant $j$ as $p_{t,j}=\frac{p_{t,j}^l+p_{t,j}^r}{2}$ where $p_{t,j}^l$ and $p_{t,j}^r$ denotes the PoG prediction on left eye image $I_{t,j}^l$ and right eye image $I_{t,j}^r$. Similarly, $p_{t,j}^l$ and $p_{t,j}^r$ are 2-dimensional vectors representing the gaze location on screen and $p_{t,j}$ averages the left and right positions.

Our InitNet follows the structure in~\cite{Park2020ECCV}. Specially, InitNet takes either the left or right eye image, parses it to ResNet18~\cite{he2015delving} and outputs $d_{t,j}^*$ for this particular image~\footnote{Note that compared to~\cite{Park2020ECCV}, we remove the pupil size prediction task as it deteriorates the gaze direction performance in practice.}. Mathematically, we denote InitNet as $f_{init}$ and $d_{t,j}^*=f_{init}(I_{t,j}^*)$, where $*=\{l,r\}$ denotes left or right sample. % $d_{t,j}^l$ is the gaze direction of the left eye image and $d_{t,j}^r$ for that of the right eye image. 
In general, $d_{t,j}^*$ is represented by a 3-dimensional unit vector and can be converted to PoG. To acheive that, we first combine $d_{t,j}^*$ and with 3D gaze origin position, which is determined during data pre-processing. This would provide us a gaze ray with 6 degree of freedom. Given camera extrinsic as well as screen plane location, we can compute the intersection of this ray with the screen plane, which gives PoG $p_{t,j}^*$. Note that theoretically, $p_{t,j}$ can be represented by pixel dimensions or in centimeters~\footnote{For instance, in EVE dataset~\cite{Park2020ECCV}, their screen is 1920 by 1080 with 553mm wide and 311mm tall.} and otherwise notified, we take the centimeter representation in our paper. %We visualize the model structure in Fig.~\ref{fig:initnet}. 
Again, InitNet is not restricted to ResNet18 but can be any State-Of-The-Art (SOTA) structure. We refer the readers to~\cite{Park2020ECCV} and Sec.~\ref{sec:exps} for more details.

~\subsubsection{Validity Module}~\label{sec:valid}
Predicting $p_{t,j}$ with %visual cues 
$I_{t,j}$ can be noisy. For instance, participants may blink eyes and given partially visible pupil, the predicted PoG can be quite off. Errors from such noisy predictions can even propagate to future samples if patterns from previous predictions are modelled.

\begin{figure}\centering
  \includegraphics[width=1.0\linewidth]{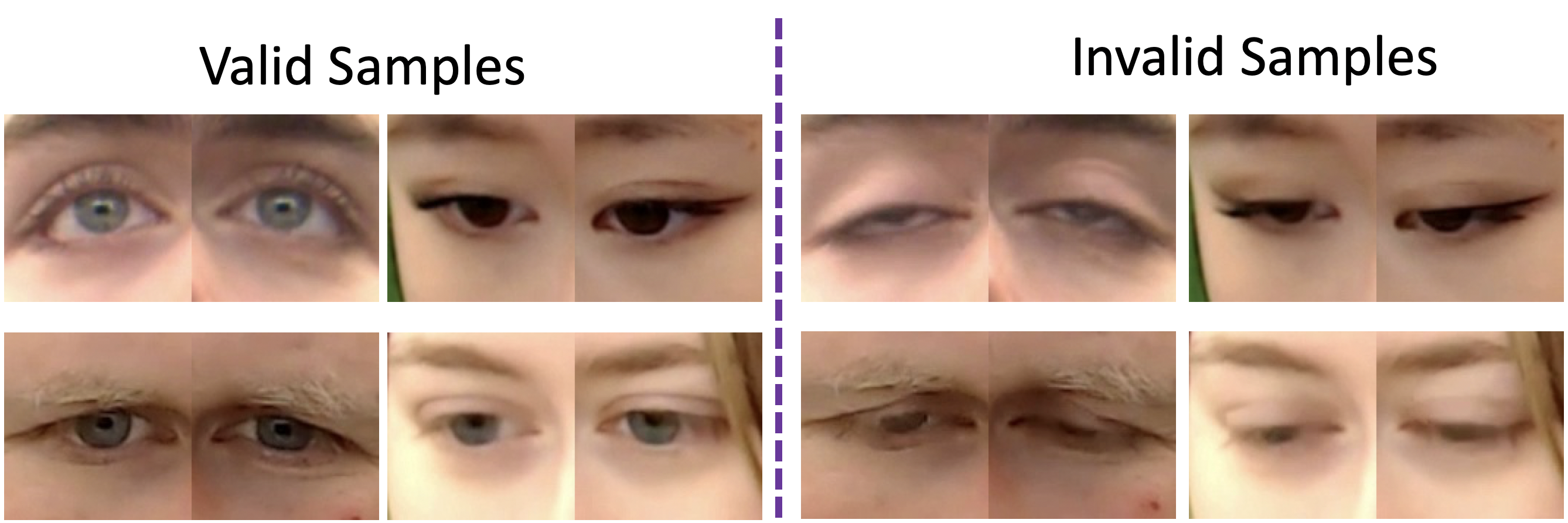}
   %\fbox{\rule{0pt}{1in} \rule{.9\linewidth}{0pt}}
  % \vspace{-0.65cm}
   \caption{
  Examples for valid and invalid samples.
   }
   \label{fig:validity_res}
\end{figure}

In view of this problem, we propose to identify how reliable $p_{t,j}$ would be and introduce our second module $f_{vm}$, or Validity Module, that takes $p_{t,j}^*$ as input and outputs $b_{t,j}^*$, where $b_{t,j}^*$ is binary denoting the reliability of current sample. %Specifically, $b_{i,j}^*$ is the probability for the reliability of $I_{i,j}^*$. 
Specifically, $f_{vm}$ is an indicate function that considers both current prediction and history predictions of the same participant, if any. $b_{t,j}^*$ is one as long as it satisfies at least one of the following two requirements. 1. The initial prediction $p_{t,j}^*$ is inside the screen. 2. $p_{t,j}^*$ lies within 3 times of the standard deviation distance w.r.t. averaged history predictions, if any%2. If we have history predictions for the $j$-th participants, for instance $\{p_{i,j}^*\}_{i<t}$, $p_{t,j}^*$ lies within 3 times of the standard deviation distance w.r.t. the average history predictions
~\footnote{We also tried to train a network to predict the $b_{t,j}^*$ with $I_{t,j}^*$ as input. In practise, the current formulation provides similar performance.}. 
Finally, $b_{t,j}=b_{t,j}^l\times b_{t,j}^r$. Example VM results are shown in Fig.~\ref{fig:validity_res}.

% In brief, our Validity Module follows ResNet18~\cite{??} structure and we simply replace the last layer with a two-class classifier. Details for learning this module can be found in Sec.~\ref{sec:model_train} and 

% Mathematically, we define our Validity Module as follows:
%\begin{equation}
%b_{i,j}= \mathbb{I} b_{i,j}^l==1 \mathbb{I} \cdot \mathbb{I} b_{i,j}^r==1 \mathbb{I}
%\end{equation}
%where $\mathbb{I} \mathbb{I}$ denotes the indicate function and $b_{i,j}^*=f_{valid}(p_{i,j}^*)$.

~\subsubsection{Self-Calibration Module}~\label{sec:calibration}
After obtaining the initial PoG $p_{t,j}$ from InitNet and reliability $b_{t,j}$ from Validity Module, our next step is to learn to compensate the prediction bias. Intuitively, every individual has a person-specific offset between his/her optical and visual axes in each eye. Although one can observe the former by the appearance of iris in $I_{t,j}$, 
%, or from $I_{t,j}$ in our setting
the later cannot be identified or detected easily as it is defined by the position of fovea at the back of eyeball. Since InitNet absorb the person-specific offset into its parameters, we further introduce Self-Calibration (SC) Module to model the offset thus to bridge the gap between visual and optical axes.

Specifically, we exploit dataset-wise distribution to model the person-specific offset.%we propose to exploit the seen gap to learn to compensate the unseen. 
\thickspace $g_{tr}$ is the average PoG on the valid samples on training set and is defined as:
\begin{equation}
    g_{tr} = \frac{\sum_{t=1,j=1}^{N,M} g_{t,j} \cdot v_{t,j}}{\sum_{t=1,j=1}^{N,M} v_{t,j}}
\end{equation}
Then we model offset by measuring the gap between $g_{tr}$ and averaged valid PoG locations and compensate our initial PoG w.r.t. learned offset to obtain the refined PoG, or $\hat{p}_{t,j}$. Mathematically, we define SC as:
\begin{equation}
\hat{p}_{t,j}= p_{t,j} - (\frac{\sum_{i \in \mathcal{H}_{t,j}} p_{i,j} \cdot b_{i,j}}{\sum_{i \in \mathcal{H}_{t,j}} b_{i,j}} - g_{tr})
\end{equation}~\label{equ:cali_general}
where $\mathcal{H}_{t,j}$ represents the set of history index. For online setting where we are given consecutive frames sequentially, $\mathcal{H}_{t,j}=\{1,\dots,t-1\}$. As for offline scenario where all samples of $j$-th participant are available, we define $\mathcal{H}_{t,j}=\{1,\dots,t-1,t+1,\dots N\}$.
%for offline scenario. 
%the online SC Module is defined as:
%\begin{equation}\label{equ:cali_online}
%\hat{p}_{t,j}^{on}= p_{t,j} - (\frac{\sum_{i=1}^{t-1} p_{i,j} \cdot b_{i,j}}{\sum_{i=1}^{t-1}b_{i,j}} - p_{tr})
%\end{equation}
%where calibration is performed w.r.t. previous seen valid images. For offline cases where all eye images of participant $j$ are available, e.g. either entire video sequences or the some independent images, we define offline SC Module as:
%\begin{equation}
%\hat{p}_{t,j}^{off}= p_{t,j} - (\frac{\sum_{i \neq t} p_{i,j} \cdot b_{i,j}}{\sum_{i \neq t}b_{i,j}} - p_{tr})
%\end{equation}~\label{equ:cali_offline}
Please note that our SC Module is person-specific since we calibrate the predictions of $p_{t,j}$ based on the information of the same participant $j$. Theoretically, this module is able to perform self-calibration as long as there are multiple samples for the same participant. More details for the two settings can be found in Sec.~\ref{sec:exps}.%, or $\sum_{i}||v_{i,j}|| \geq 2$. 
% This module is capable of handling cross-person test where test participants are different from these involved in training as long as they satisfied our multiple-sample requirement. 

\begin{figure}\centering
  \includegraphics[width=1.0\linewidth]{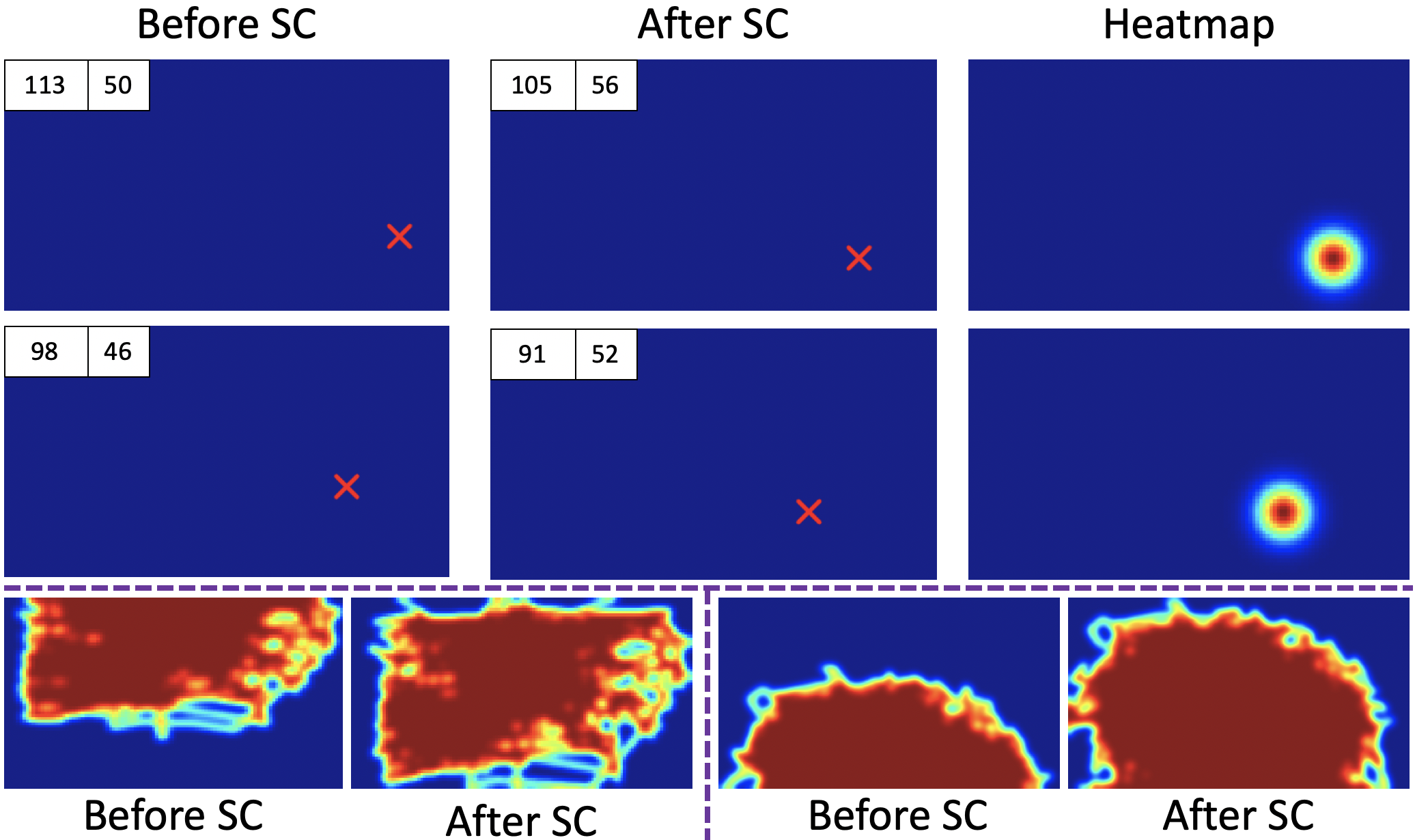}
  %\fbox{\rule{0pt}{1in} \rule{.9\linewidth}{0pt}}
  % \vspace{-0.65cm}
  \caption{
  Example results of SC Module. SC is able to perform translation on initial predictions w.r.t. valid initial predictions and dataset-wise distribution.% (top right). 
  }
  \label{fig:calibration_res}
\end{figure}

We denote the refined PoG predictions for the $t$-th sample of the $j$-th participant obtained from SC Module as $\hat{p}_{t,j}$. 
%and omit $on$ or $off$ for clearness. 
The prediction history of this sample is defined as $\mathcal{P}_{t,j} = \{\hat{p}_{i,j}\}_{i \in \mathcal{H}_{t,j}}$. We visualize the input $p_{t,j}$ and output of SC Module $\hat{p}_{t,j}$ in Fig.~\ref{fig:calibration_res} (See Before SC and After SC on the upper region). For clarity, the predicted PoG (in pixel space) are also reported by the 2D vector inside each image.

~\subsubsection{Person-specific Transform Module}~\label{sec:calibration}
Although SC Module is beneficial in terms of removing the person-specific difference, it only deals with cases where the difference is mainly translation. To this end, we further propose the Person-specific Transform (PT) Module to model and compensate more general differences.

Specifically, our PT Module takes the heatmaps of the refined PoG prediction $\hat{p}_{t,j}$ and its history $\mathcal{P}_{t,j}$ as input and outputs the transformed heatmaps for both. To obtain the heatmap for the $t$-th sample of $j$-th participant, or $r_{t,j}\in \mathbb{R}^{H \times W \times 1}$, we first map $\hat{p}_{t,j}$ to pixel space and then convolving a dirac delta function centered at $\hat{p}_{t,j}$ with a isotropic 2D Gaussian of fixed variance (See the middle and right column in Fig.~\ref{fig:calibration_res} for $\hat{p}_{t,j}$ and $r_{t,j}$). We further denote the heatmap for $\mathcal{P}_{t,j}$ as $r_{t,j}^h \in \mathbb{R}^{H \times W \times 1}$. To obtain it, we repeat similar process described above on only valid samples. More specifically, $\hat{p}_{i,j}$ is valid as long as $\hat{p}_{i,j} \in \mathcal{P}_{t,j}$ and $b_{i,j} \neq 0$. Details of the process can be found in Alg.~\ref{alg:algorithm1_generate_history_heatmap}.

In practice, we find the order of samples does not affect the final performance. More importantly, instead of simply combining the heatmap of all individual valid points in $\mathcal{P}_{t,j}$ to obtain the history heatmap, we propose to include the $trajectories$ between two PoGs as well, as explained in the 11-th step in Alg.~\ref{alg:algorithm1_generate_history_heatmap}. Such design helps us to focus more on the PoGs that are distributed far away from center so that they can play more important role in our modelling process. %In addition, we would like to 
Please note that we exclude invalid PoGs in this generation process in order to perform meaningful transformation, e.g. our PT Module will not be distracted by invalid predictions.

We concatenate $r_{t,j}$ and $r_{t,j}^h$ channel-wise to obtain our input~$\textbf{r}_{t,j}\in \mathbb{R}^{H \times W \times 2}$ for PT Module. Our PT Module consists of one localization net, one grid generator and one sampler~\cite{NIPS2015_33ceb07b}. The output of PT Module is denoted as $\hat{\textbf{r}}_{t,j}\in\mathbb{R}^{H \times W \times 2}$, including both transformed sample heatmap $\hat{r}_{t,j}$ and transformed history heatmap $\hat{r}_{t,j}^h$. Both of them are in $\mathbb{R}^{H \times W \times 1}$ space. Mathematically, we denote PT Module as $f_{pt}$ and $\hat{\textbf{r}}_{t,j}=f_{pt}(\textbf{r}_{t,j})$. To obtain the final PoG for $I_{t,j}$, one just need to apply softmax on $~\hat{r}_{t,j}$, find the position with the maximum value and then re-scale the the position to screen size. We give some example results of PT in Fig.~\ref{fig:stt_res} (PT Inference).
More details of our PT Module can be found in supplementary materials. 
\begin{figure}\centering
  \includegraphics[width=1.0\linewidth]{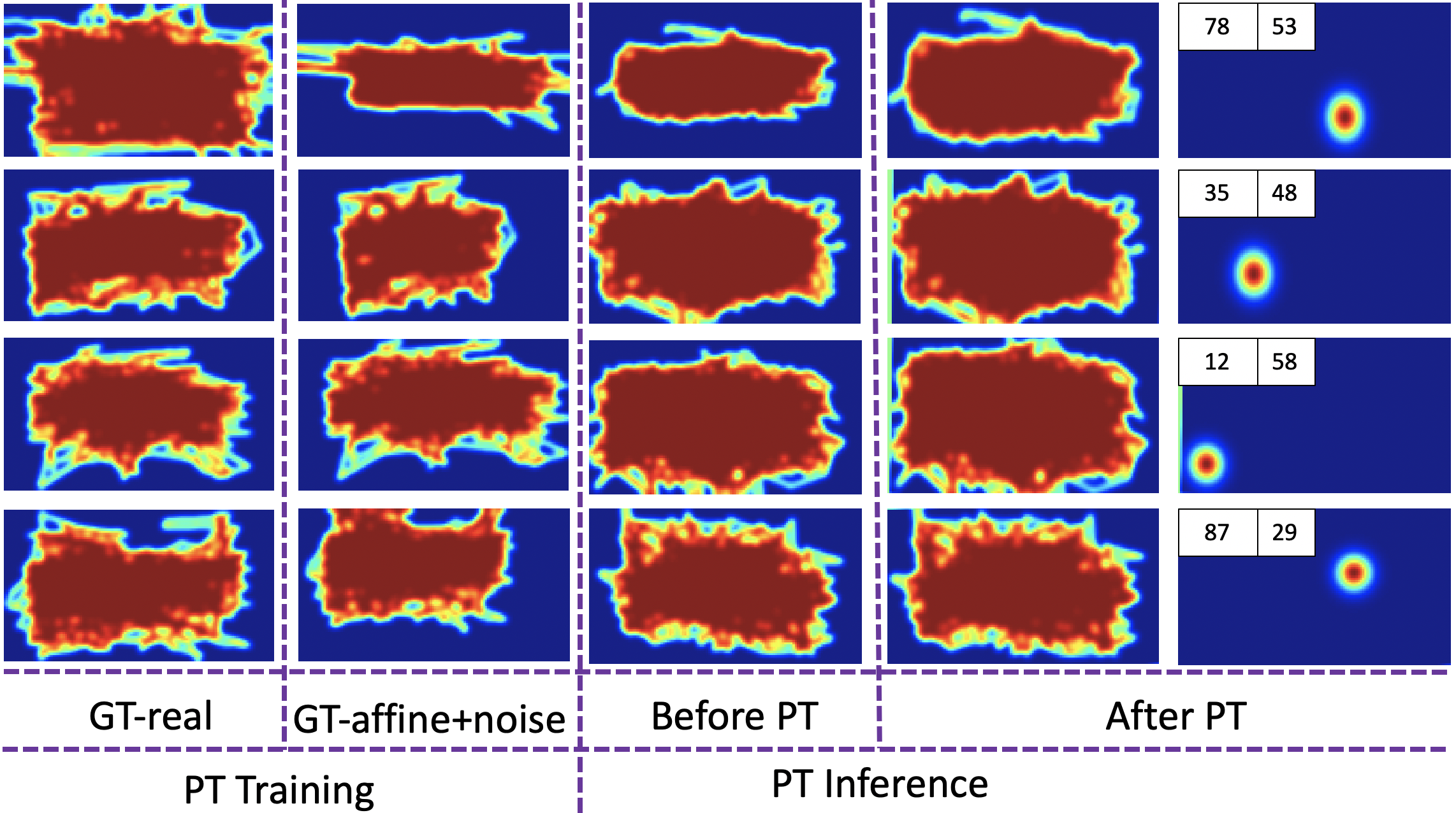}
  %\fbox{\rule{0pt}{1in} \rule{.9\linewidth}{0pt}}
  % \vspace{-0.65cm}
  \caption{
  Examples for PT Module. We introduce diverse affine-like transformation with noise during training (left), which enables us to handle unseen cases thus we can effectively model cross-person differences during inference (right). We omit $r_{t,j}$ in Before PT.% (top right). 
  }
  \label{fig:stt_res}
\end{figure}

Note that SC Module is necessary yet important even SC seems to only considers translation-like offset while PT can compensate all affine-like transformations. For instance, PT is not able to handle scenarios where initial PoG predictions $p_{i,j}$ are mostly off screen. Or in another word, the translation is so large that some samples are out of scope. SC Module, in this case, is able to translate the initial predicted PoGs to some reasonable locations so that the later learning process of PT is meaningful and effective. In general, SC Module and PT Module are mutual beneficial and play different roles in our full model. We refer the readers to PT training in Fig.~\ref{fig:stt_res} and lower part of Fig.~\ref{fig:calibration_res} for different roles that SC and PT play in our model.

~\subsection{Model Training}~\label{sec:model_train}
We demonstrate our full model in previous section with the assumption that supervisions are all available for each module. In this section, we will describe how to train our model with given $~\mathcal{D}$.
~\paragraph{InitNet Module} Given the structure defined in Sec.~\ref{sec:eyenet}, we can prepare the ground-truth gaze direction $\widetilde{d}_{t,j}$ with corresponding $g_{t,j}$ for each $I_{t,j}$. Then our loss function is defined as:
\begin{equation}\label{equ:cali_online}
\mathcal{L}_{init}(\widetilde{d}_{t,j},d_{t,j})=\frac{180}{\pi}\arccos(\frac{\widetilde{d}_{t,j}\cdot d_{t,j}}{||\widetilde{d}_{t,j}|| ||d_{t,j}||})
\end{equation}
%~\paragraph{Validity Module}
%Similarly, with given binary $v_{t,j}$, we can train our model with the following loss function $\mathcal{L}_{valid}(v_{t,j},b_{t,j})$ as the BECLoss between prediction and ground-truth.
~\paragraph{Person-specific Transform Module}
The PT Module is designed to remove the remaining person-specific difference in $\textbf{r}_{t,j}$ regardless of which network structure is used for the Initnet Module. As the characteristics of person-specific difference reflected in $\textbf{r}_{t,j}$ varies across datasets and Initnet Modules, a generalised PT Module should be trained on samples that cover all these variations. In practice, we augment the ground-truth $\widetilde{\textbf{r}}_{t,j}$ to $\check{\textbf{r}}_{t,j}$ in order to emulate all variations in person-specific difference. During training, $\check{\textbf{r}}_{t,j}$ is served as input to the PT Module and $\widetilde{\textbf{r}}_{t,j}$ as ground truth; during inference, the PT Module takes in $\textbf{r}_{t,j}$ and outputs $\hat{\textbf{r}}_{t,j}$. $\widetilde{\textbf{r}}_{t,j}$ is obtained by concatenate $\widetilde{r}_{t,j}$ and $\widetilde{r}_{t,j}^h$ channel-wise and similarly $\check{\textbf{r}}_{t,j}$ is obtained by concatenate $\check{r}_{t,j}$ and $\check{r}_{t,j}^h$ channel-wise. Similar to the process of obtaining $r_{t,j}$ and $r_{t,j}^h$ from $\hat{p}_{t,j}$, $b_{i,j}$ and $\mathcal{H}_{t,j}$ as shown in Alg.~\ref{alg:algorithm1_generate_history_heatmap}, we obtain $\widetilde{r}_{t,j}$ and $\widetilde{r}_{t,j}^h$ from $g_{i,j}$, $v_{i, j}$ and $\mathcal{H}_{t,j}$, and obtain $\check{r}_{t,j}$ and $\check{r}_{t,j}^h$ from $\check{g}_{i,j}$, $v_{i, j}$ and $\mathcal{H}_{t,j}$, where $\check{g}_{i,j}$ is augmented from $g_{i,j}$ as shown in Alg.~\ref{alg:algorithm2_augmentation}. We refer the readers to supplementary for details on how we define the set for affine parameters $\mathcal{A}$ and random noise $\mathcal{N}$ during this augmentation process. We provide some visualization examples for $\widetilde{r}_{t,j}^h$ and $\check{r}_{t,j}^h$ in PT training part of Fig.~\ref{fig:stt_res}, where $GT$-$real$ denotes the ground-truth heatmap and $GT$-$affine+noise$ is the augmented one that emulates person-specific difference.

As long as we have training samples available for PT Module, we then define our loss function as:
\begin{equation}
    \mathcal{L}_{pt}(\widetilde{\textbf{r}}_{t,j},\check{\textbf{r}}_{t,j})=BCEloss(\widetilde{r}_{t,j}^h,\check{r}_{t,j}^h)
\end{equation}
where BCE is the binary cross-entropy loss. As can be seen in the equation above, we do not introduce any loss on transformed $\check{r}_{t,j}$. Our PT Module learns a transformation w.r.t. $\check{r}_{t,j}^h$ and applies the learned transformation directly on $\check{r}_{t,j}$. Although one can always introduce loss on $\check{r}_{t,j}$, in practice, we find that introducing either BCEloss on $\check{r}_{t,j}$ or MSE loss on final numerical estimate of PoG deteriorates the model performance. This might because that learning transformation with per-sample gaze heatmap $\check{r}_{t,j}$ is less meaningful as there are only limited information on this heatmap. In addition, focusing on $\check{r}_{t,j}$ might further confuse the localization net in PT Module.
% , which provides $\hat{r}_{t,j}$

% $\mathcal{P}_{t,j} = \{\hat{p}_{i,j}\}_{i \in \mathcal{H}_{t,j}}$

\begin{algorithm}
\caption{Framework for generating history heatmap ${r}_{t,j}^h$.}
	\label{alg:algorithm1_generate_history_heatmap}
	\begin{algorithmic}[1]
	\STATE \textbf{Input:} 
	Predicted history samples: $\mathcal{P}_{t,j} = \{\hat{p}_{i,j}\}_{i \in \mathcal{H}_{t,j}}$; Validity prediction: $\{b_{i,j}\}$ for all $i\in\mathcal{H}_{t,j}$
	\STATE \textbf{Output:} Heatmap $r_{t,j}^h$ for predicted history samples;
	\STATE \textbf{Initialize:} Set of visited samples: $\mathcal{S}=\mathcal{H}_{t,j}$; All zero heatmap: $r_{t,j}^h$
	\STATE Randomly select $k$ from $\mathcal{S}$ as long as $ b_{k,j} \neq 0$;
	\STATE Map $\hat{p}_{k,j}$ to pixel space and plot in $r_{t,j}^h$ by setting the pixel value of corresponding location to 1.
	\STATE Update: $\mathcal{S}=\mathcal{S}\nsubseteq k$ 
	\WHILE{$\sum_{i\in\mathcal{S}} b_{i,j}$ greater than 0} 
	\STATE Randomly select $m$ from $\mathcal{S}$ as long as $b_{m,j} \neq 0$
	\STATE Map $\hat{p}_{m,j}$ to pixel space and plot it in $r_{t,j}^h$ 
	\STATE Update: $\mathcal{S}=\mathcal{S}\nsubseteq m$ 
	\STATE In $r_{t,j}^h$, set the value of all pixels that lie in the segment bounded with two points, or $\hat{p}_{m,j}$ and $\hat{p}_{k,j}$, to 1 
	\STATE Set $k$ to $m$ 
	\ENDWHILE
    \STATE $r_{t,j}^h=\textit{g}(r_{t,j}^h)$ where $\textit{g}$ denotes 2D Gaussian of fixed variance.
    \end{algorithmic}
\end{algorithm}

\begin{algorithm}
\caption{Framework for generating the augmented samples $\check{g}_{i,j}$ from $g_{i,j}$.}
	\label{alg:algorithm2_augmentation}
	\begin{algorithmic}[1]
	\STATE \textbf{Input:} Ground truth samples: $\{g_{i,j}\}$;
	\STATE \textbf{Output:} Augmented samples $\{\check{{g}_{i,j}}\}$ 
	\STATE \textbf{Initialize:} Affine transformation parameters:$\mathcal{A}$; Noise parameter:$\mathcal{N}$
	\STATE Randomly sample one parameter set $a$ from $\mathcal{A}$
	
	\FOR{$g_{m,j}\in\{g_{i,j}\}$} 
	\STATE Initialize: $\check{g}_{m,j}$ = $g_{m,j}$
	\STATE Randomly sample one parameter set $n_{m}$ from $\mathcal{N}$
	\STATE Map $\check{g}_{m,j}$ to pixel space
	\STATE Apply affine transformation $a$ to $\check{g}_{m,j}$ in pixel space, add $n_m$ on top of the transformed location
	\STATE Convert $\check{g}_{m,j}$ in pixel space back to direction space
	\ENDFOR
    \end{algorithmic}
\end{algorithm}

%% file: sections/experiments.tex
~\section{Experiments}~\label{sec:exps}
\begin{table*}
  \centering\small
  \input{tables/eve_main_result}
  \caption{Performances on EVE test set. We outperform the SOTA significantly when having the same setting. And our method beats the SOTA with less information required. We rank the first in EVE leader-board upon submission.}
  \label{tbl:eve_res_main}
\end{table*}

In this section, we demonstrate the effectiveness of our proposed model by conducting several experiments on three publicly available datasets, EVE~\cite{Park2020ECCV}, MPIIGaze~\cite{zhang2017mpiigaze} and XGaze~\cite{zhang2020eth}. We demonstrate the state-of-the-art (SOTA) performance on these datasets and perform ablation study by validating the effectiveness of each module.
\vspace{-0.3cm}
\paragraph{Datasets:}\textbf{EVE} dataset is the main one that validate our ideas on. Specifically, 12,308,334 frames are provided in EVE. 54 participants are recorded with natural eye movements (as opposed to following specific instructions or smoothly moving targets). The gaze angles in EVE are in the range of [-60,60] degrees in the vertical and [-70, +70] degrees in the horizontal direction. It also covers a large set of head movement. We follow the standard split in~\cite{Park2020ECCV} and report our final results on test sequence. We conduct the ablation study on its validation set.
~\textbf{MPIIGaze} consists of 213,659 images from 15 participants (six females, five with glasses), among which 14 of them are used for training and one for testing. In addition, 10 participants had brown, 4 green, and one grey iris colour. Participants collected the data over different time periods thus images are with high illumination diversity. The gaze angles in MPIIGaze are in the range of [-1.5,20] degrees in the vertical and [-18, +18] degrees in the horizontal direction. We follow the standard split in~\cite{zhang2017mpiigaze} and report our results.
~\textbf{XGaze} dataset consists of 1,083,492 images taken from 110 participants, including 47 female and 63 male. 17 of them wore contact lenses and 17 of them wore eyeglasses during recording. As for illumination condition, 16 controlled conditions are explored in XGaze. The gaze angles in XGaze are in the range of [-70,70] degrees in the vertical and [-120, +120] degrees in the horizontal direction. We follow the standard split as suggested in~\cite{zhang2020eth}, e.g. with-in dataset setting where 80 participants are used for training and 15 are for testing, and report our performance on test set.
\vspace{-0.3cm}
\paragraph{Evaluation Metrics:} As for EVE, We follow~\cite{Park2020ECCV} and report the predicting gaze direction and PoG. For the~\cite{zhang2020eth,zhang2017mpiigaze}, we report the gaze prediction error in degree.
\vspace{-0.3cm}
\paragraph{Training Details:} As for EVE dataset, we train InitNet Module from scratch for 8 epochs with ADAM~\cite{kingma2014adam}, with learning rate set to $1e-3$. Person-specific Transform (PT) Module is trained from scratch with SGD for 100 epochs and we set learning rate to 0.1 with momentum set to 0.9. Batch size is set to be 12 and 3200 for InitNet and PT respectively. As for MPIIGaze and XGaze datasets, we exploit~\cite{zhang2017mpiigaze} and~\cite{zhang2020eth} structure instead to demonstrate the generality of proposed method, e.g. our method is not restricted to any specific gaze prediction network but is able to improve the performance of existing methods in general. Please further note that we train our PT Module~\textit{only} on EVE and directly apply the pre-trained model on MPIIGaze and XGaze. Also, since both MPIIGaze and XGaze include only valid frames/images, we do not apply Validity Module to them. We have both online and offline version for EVE dataset as it provides video sequences. Only offline model is available on MPIIGaze and XGaze due to the lack of consecutive frames. We set $H$ to 72 and $W$ to 128 in experiment.

\begin{figure}\centering
  \includegraphics[width=1.0\linewidth]{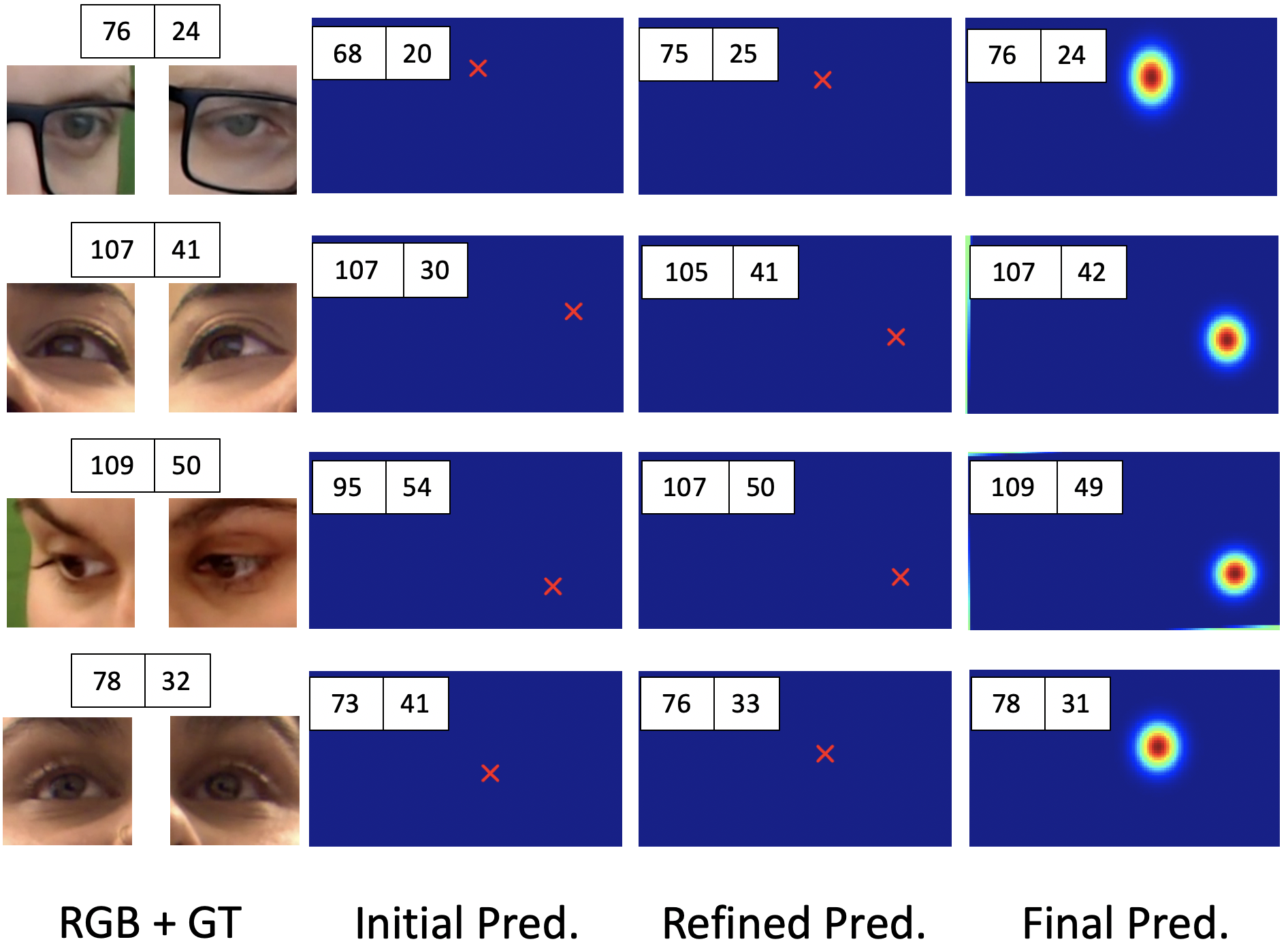}
  %\fbox{\rule{0pt}{2in} \rule{.9\linewidth}{0pt}}
  % \vspace{-0.65cm}
  \caption{
  We demonstrate input RGB, GT, initial, refined and final predictions in this figure. Introducing each module can indeed boost performance.
  }
  \label{fig:final_results_eve}
\end{figure}

\subsection{Results on EVE}
In this section, we evaluate our method on EVE and demonstrate that our proposed method can aid in gaze estimation. Developed upon InitNet, we then evaluate the effects of individual modules in improving initial estimate of PoGs. 
\vspace{-0.3cm}
\subsubsection{Main Results}
\paragraph{Eye Gaze Estimation} We first consider the task of gaze estimation purely from paired eye image patches. To obtain the ground-truth, we simply average the ground-truth of left and right eye and compared the averaged ground-truth with our prediction. Tab.~\ref{tbl:eve_res_main} shows the performance of our full model on predicting gaze direction and PoG. We can see that our proposed method beats SOTA significantly on EVE, $27.6\%$ relative improvement. Please note that the proposed method is directly comparable to SOTA~\cite{Park2020ECCV} as all networks are trained on the training split of EVE. To have a fair comparison, our InitNet follows the same structure as EyeNet~\cite{Park2020ECCV} (See Sec.~\ref{sec:method} for details). More importantly, in contrast to incorporating contents in screen or requiring consecutive frames as input, our model relies only on eye image patches thus is more general and requires less information. Generally, we find that modelling person-specific differences is very important and our proposed modules, SC and PT, indeed benefits a lot in terms of compensate for the person-specific pattern.
\vspace{-0.3cm}
\paragraph{Qualitative Results}
We visualize our results qualitatively in Fig.~\ref{fig:final_results_eve}. We can see that when provided with initial estimates of PoG from InitNet, our Self-Calibration (SC) Module nicely recovers person-specific offsets at test time to yield improved estimates of PoG. Introducing PT Module further boosts the performance by compensating more general yet diverse offsets. By comparing the Ground Truth (GT) with our step-wise results, the mutual beneficial of SC and PT Modules are more observable. Compared to SOTA~\cite{Park2020ECCV} that requires additional screen contents, our method is more general and is applicable to various datasets.
\vspace{-0.3cm}
\subsubsection{Ablation Study}
To demonstrate the effectiveness of each module we introduced, we further conduct experiments on incrementally adding modules. We validate our ideas on EVE validation set and report the quantitative number in Tab.~\ref{tbl:eve_res_ablation}. As can be seen in this table, each module is indeed beneficial for PoG prediction task and combining all of them gives the best performance. Moreover, we are also able to beat the SOTA~\cite{Park2020ECCV} method that requires both consecutive frames and screen contents as input on EVE validation set. We also perform ablation studies on the impact of history length, and we refer the readers to supplementary for more details.

\begin{table}[t!]
  \centering\small
  \input{tables/eve_ablation_res}
  \caption{Ablation study on EVE validation set. Each component is beneficial for our task and our method outperforms the SOTA by a large margin. %In addition, we can see that introducing each module is beneficial for gaze prediction task.
  }
  \label{tbl:eve_res_ablation}
\end{table}

\subsection{Results on MPIIGaze and XGaze}
We conduct experiments on two more datasets, MPIIGaze and XGaze. Please note that to have fair comparisons as well as to demonstrate that our method is not restricted to specific network structure, we replace our InitNet structure with~\cite{zhang2017mpiigaze} and~\cite{zhang2020eth} for MPIIGaze and XGaze, respectively. Moreover, to showcase the generality of our proposed method, we train PT Module on neither of these two datasets, but directly applying PT Module trained on EVE and report the performance in Tab.~\ref{tbl:additional_res}. As can be seen in this table, the proposed method, again, outperform the SOTA by a margin, or relatively $32.9\%$ improvement on MPIIGaze and $36.0\%$ improvements on XGaze. Our results suggest that our proposed method can indeed model the person-specific difference very well, even on cross dataset evaluation. More importantly, our proposed method is general enough to improve various existing models and able to achieve best results. 

\begin{figure}\centering
  \includegraphics[width=1.0\linewidth]{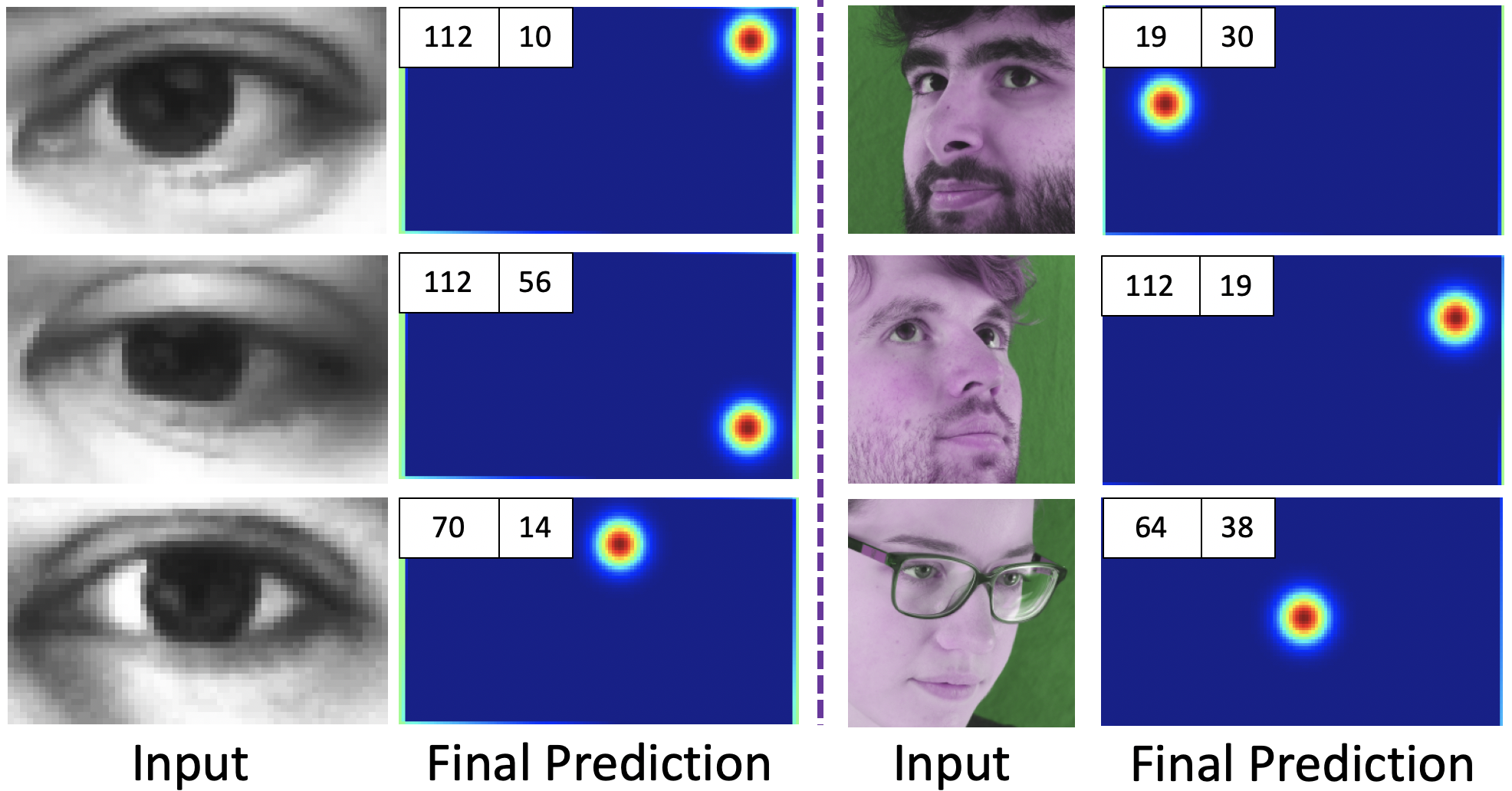}
  %\fbox{\rule{0pt}{2in} \rule{.9\linewidth}{0pt}}
  % \vspace{-0.65cm}
  \caption{
  We visualize examples for MPIIGaze and XGaze on the left and right respectively. 
  }
  \label{fig:final_results_others}
\end{figure}

We also visualize our results on MPIIGaze and XGaze in Fig.~\ref{fig:final_results_others}. As can be seen in these figure, our model can handle various type of input and can provide visually satisfactory gaze predictions in both datasets.

\begin{table}[t!]
  \centering\small
  \input{tables/mpii_xgaze}
  \caption{Performances on MPIIGaze and XGaze. Again, we outperform the SOTA significantly. %Please note that we train PT Module on neither of them but instead applying PT that pre-trained on EVE and report the performance.
  }
  \label{tbl:additional_res}
\end{table}

\vspace{-0.7cm}

%% file: tables/eve_main_result.tex
\begin{tabular}{l|c|c|c|ccc}
  \hline
              &
              \multicolumn{3}{c|}{Requirements} &\multicolumn{3}{c}{Performance on EVE test set~\cite{Park2020ECCV}} \\
  \hline
  Method      & Static Image & Consecutive Frame & Screen Content & Gaze Dir. ($^{\circ}$) $\downarrow$ & PoG (cm) $\downarrow$ & PoG (px)  $\downarrow$\\
  \hline
  EyeNet-Static~\cite{Park2020ECCV}  & \checkmark & & & 4.54 &5.10 &172.7 \\
  \hline
  EyeNet-GRU~\cite{Park2020ECCV}  & & \checkmark & & 3.48 & 3.85 & 132.56 \\
  GazeRefineNet-Static~\cite{Park2020ECCV}  & & \checkmark & \checkmark & 2.87 & 3.16 & 109.85 \\
  GazeRefineNet-RNN~\cite{Park2020ECCV}  & & \checkmark & \checkmark & 2.57 & 2.83& 98.38
  \\
  GazeRefineNet-LSTM~\cite{Park2020ECCV}  & & \checkmark & \checkmark & 2.53 &2.79& 96.97 \\
  GazeRefineNet-GRU~\cite{Park2020ECCV}  & & \checkmark & \checkmark & 2.49 & 2.75& 95.59 \\
  \hline
  Ours-Online & & \checkmark & & 2.15 & 2.39 & 82.91 \\
  Ours-Offline & \checkmark & & &\textbf{1.95} & \textbf{2.17}& \textbf{75.19}  \\
  \hline
\end{tabular}

%% file: tables/eve_ablation_res.tex
\begin{tabular}{l|ccc}
  \hline
             \multicolumn{4}{c}{Performance on EVE validation set~\cite{Park2020ECCV}} \\
  \hline
  Method     &Gaze Dir. ($^{\circ}$) $\downarrow$ & PoG (cm) $\downarrow$ & PoG (px)  $\downarrow$\\
  \hline
  ~\cite{Park2020ECCV} & 2.1 & - & - \\  
  \hline
  InitNet & 2.41 & 2.72 & 94.33 \\
  \hline
  +SC & 2.32 & 2.66 & 92.19 \\
  +SC+VM & 2.26 & 2.57 & 89.39 \\
  Full(online)& 2.04 & 2.33 & 80.74 \\
  Full(offline)& \textbf{1.89} & \textbf{2.16} & \textbf{74.85}\\  
  \hline 
\end{tabular}

%% file: tables/mpii_xgaze.tex
\begin{tabular}{l|cc}
  \hline
             & MPII test set~\cite{zhang2017mpiigaze} & XGaze test set~\cite{zhang2020eth} \\
  \hline
  Method     &Gaze Dir. ($^{\circ}$) $\downarrow$ & Gaze Dir. ($^{\circ}$) $\downarrow$\\
  \hline
  ~\cite{zhang2020eth} & 4.8 & 4.5 \\  
  ~\cite{park2019few} & 5.2 & - \\  
  ~\cite{fischer2018rt} & 4.8 & - \\  
  ~\cite{park2018deep} & 4.5 & - \\  
  \hline
  InitNet & 5.73/4.83 & 4.50 \\
  Ours-Offline & 4.14/\textbf{3.02} & \textbf{2.88} \\
  \hline 
\end{tabular}

%% file: sections/conclusion.tex
~\section{Conclusion}~\label{sec:conclusion}
In this paper, we propose a novel method that aims to improve the cross-person gaze prediction task where only eye images are given as input and Point of Gaze (PoG) is the output. To this end, we introduce Validity Module (VM) to handle noisy data and two other modules, SC and PT, to explicitly model person-specific differences in PoG prediction task. We estimate the reliability of each sample in VM and considers only valid ones for later difference modelling process. Our SC considers mainly the translation-like offsets and we further introduce PT to take into account more general and person-specific affine-like differences. We demonstrate the effectiveness of our proposed method on three publicly available datasets and report the SOTA performance. We also showcase the effectiveness and usefulness of each module in our ablation study.